\documentclass{article}
\usepackage{nips13submit_e,times}
\usepackage{graphicx}
\usepackage{amsmath}
\usepackage{amssymb}
\usepackage{comment}

\title{Unsupervised Feature Learning by Deep Sparse Coding}
\author{Yunlong He\thanks{heyunlong@gatech.edu, Georgia Institute of Technology}
\And
Koray Kavukcuoglu\thanks{koray@deepmind.com, DeepMind Technologies}
\And Yun Wang\thanks{yunwang@princeton.edu, Princeton University}
\And Arthur Szlam \thanks{aszlam@ccny.cuny.edu ,The City College of New York}
\And Yanjun Qi\thanks{yanjun@virginia.edu, University of Virginia}
}
\nipsfinalcopy
\begin{document}
\maketitle

\vspace{-2mm}
\begin{abstract} \small\baselineskip=9pt  In this paper, we propose a new unsupervised feature learning framework, namely Deep Sparse Coding (DeepSC), that extends sparse coding to a multi-layer architecture for visual object recognition tasks. The main innovation of the framework is that it connects the sparse-encoders from different layers by a sparse-to-dense module. The sparse-to-dense module is a composition of a local spatial pooling step and a low-dimensional embedding process, which takes advantage of the spatial smoothness information in the image. As a result, the new method is able to learn several levels of sparse representation of the image which capture features at a variety of abstraction levels and simultaneously preserve the spatial smoothness between the neighboring image patches. Combining the feature representations from multiple layers, DeepSC achieves the state-of-the-art performance on multiple object recognition tasks.
\end{abstract}

\vspace{-2mm}
\section{Introduction}
\vspace{-2mm}
Visual object recognition is a major topic in computer vision and machine learning. In the past decade, people have realized that the central problem of object recognition is to learn meaningful representations (features) of the image/videos. A large amount of focus has been put on constructing effective learning architecture that combines modern machine learning methods and in the meanwhile considers the characteristics of image data and vision problems.

In this work, we combine the power of deep learning architecture and the bag-of-visual-words (BoV) pipeline to construct a new unsupervised feature learning architecture for  learning image representations.  Compared to the single-layer sparse coding (SC) framework, our method can extract feature hierarchies at the different levels of abstraction. The sparse codes at the same layer keeps the spatial smoothness across image patches and different SC hierarchies also capture different spatial scopes of the representation abstraction.  As a result, the method has richer representation power and hence has better performance on object recognition tasks. 
 Compared to deep learning methods, our method benefits from effective hand-crafted features, such as SIFT features, as the input. Each module of our architecture has sound explanation and can be formulated as explicit optimization problems with promising computational performance. The method shows superior performance over the state-of-the-art methods in multiple experiments.

In the rest of this section, we review the technical background of the new framework, including the pipeline of using bag-of-visual-words for object recognition and a low-dimensional embedding method called DRLIM.

\subsection{Bag-of-visual-words pipeline for object recognition}
\label{sc}

\begin{figure*}[!]
\centerline{\includegraphics[width=5.5in]{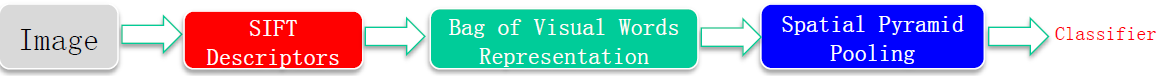}}
\caption{The bag-of-visual-words pipeline.}\label{arch1}
\end{figure*}
We now review the bag-of-visual-words pipeline consisting of hand-crafted descriptor computing, bag-of-visual-words representation learning, spatial pyramid pooling and finally a classifier. 

The first step of the pipeline is to exact a set of overlapped image patches from each image with fixed patch size, while the spacing between the centers of two adjacent image patches is also fixed. Then a $D$-dimensional hand-crafted feature descriptor (e.g. $128$-dimensional SIFT descriptor) is computed from each image patch. Now let $X^{(i)}$ denote the set of $M_i$ feature descriptors, which are converted from $M_i$ overlapped image patches extracted from the $i$-th image (e.g. size $300 \times 300$), i.e.,
$$X^{(i)}=[x^{(i)}_1,\cdots, x^{(i)}_{M_i}] \in \mathbb R^{D \times M_i},$$
where $x^{(i)}_j$ is the feature descriptor of the $j$-th patch in the $i$-th image.

Let $X=[X^{(1)}, X^{(2)}\cdots, X^{(N)}] \in \mathbb R^{D \times M}$, where $M=M_1+M_2+\cdots+M_N$, denote the set of all feature descriptors from all $N$ training images. The second step of the pipeline consists of a dictionary learning process and a bag-of-visual-words representation learning process. In the case of using sparse coding to learn the bag-of-visual-words representation, the two processes can be unified as the following problem.
\begin{align}\label{dictlearn}
\min_{V, Y} & \|X-VY\|_F^2+\alpha \|Y\|_{1,1} \\
 \notag =&\sum_{m=1}^{M}\|x_m-Vy_m\|_2^2+\alpha \|y_m\|_1 \\
s.t. & \|v_k\|\le 1, \ \ \forall k=1, \cdots, K \notag
\end{align}
where $V=[v_1, \cdots, v_K] \in \mathbb R^{D \times K}$ denotes the dictionary of visual-words, and columns of $Y=[y_1,\cdots, y_{M}] \in \mathbb R^{K \times M}$ are the learned sparse codes, and $\alpha$ is the parameter that controls sparsity of the code. We should note, however, other sparse encoding methods such as vector quantization and LLC could be used to learn the sparse representations (see~\cite{coates2011importance} for review and comparisons). Moreover, the dictionary learning process of finding $V$ in \eqref{dictlearn} is often conducted in an online style~\cite{mairal2009online} and then the feature descriptors of the $i$-th image stored in $X^{(i)}$ are encoded as the  bag-of-visual-words representations stored in $Y^{(i)}=[y^{(i)}_1,\cdots, y^{(i)}_{M_i}] $ in the $K$-dimensional space ($K>>D$). Intuitively speaking, the components of the bag-of-visual-words representation are less correlated compared to the components of dense descriptors. Therefore, compared to the dense feature descriptors, the high-dimensional sparse representations are more favorable for the classification tasks.

In the third stage of the pipeline, the sparse bag-of-visual-words representations of all image patches from each image are pooled together to obtain a single feature vector for the image based on the histogram statistics of the visual-words. To achieve this, each image is divided into three levels of pooling regions as suggested by the spatial pyramid matching (SPM) technique~\cite{lazebnik2006beyond}.
The first level of pooling region is the whole image. The second level is consist of 4 pooling regions which are 4 quadrants of the whole image.  The third level consist of 16 pool regions which are quadrants of the second level pooling regions. In this way, we obtain 21 overlapped pooling regions. Then for each pooling region, a max-pooling operator is applied to all the sparse codes whose associating image patch center locates in this pooling region, and we obtain a single feature vector as the result. The max-pooling operator maps any number of vectors that have the same dimensionality to a single vector, whose components are the maximum value of the corresponding components in the mapped vectors. Formally, given the descriptors $y_1, \cdots, y_n \in \mathbb R^{K}$ that are in the same pooling region, we calculate 
\begin{equation}
\label{max}
y=op_{max}(y_1, \cdots, y_n): = max\{y_1, \cdots, y_n\} \in \mathbb R^{K},
\end{equation} where max is operated component-wisely.  From the second stage of the framework, we know that the nonzero elements in a sparse code imply the appearance of corresponding visual-words in the image patch. Therefore, the max-pooling operator is actually equivalent to calculating the histogram statistics of the visual-words in a pooling region. Finally,  the pooled  bag-of-visual-words representations from 21 pooling regions are concatenated to obtain a single feature vector, which is regarded as the representation for the image and linear SVM is then used for training and testing on top of this representation. Since the labels of the training images are not used until the final training of SVM, the whole pipeline is regarded as an unsupervised method. For the rest of this paper, we focus on the version of the pipeline where the feature (bag-of-visual-words representation) learning part is performed by a sparse coding step as in \eqref{dictlearn}. 

\subsection{Dimensionality reduction by learning an invariant mapping} \label{drlim}

We now review a method called dimensionality reduction by learning an invariant mapping (DRLIM, see~\cite{hadsell2006dimensionality}), which is the base model for our new method in Subsection \ref{sec-dr}. Different from traditional unsupervised dimensionality reduction methods, DRLIM relies not only on a set of training instances $y_1, y_2, \cdots,  y_n  \in \mathbb R^K$, but also on a set of binary labels  $\{l_{ij}: (i,j) \in I\}$, where $I$ is the set of index pairs such that $(i,j)\in I$ if the label for the corresponding instance pair $(y_i, y_j)$ is available. The binary label $l_{ij}=0$ if the pair of training instances $y_i$ and $y_j$ are similar instances, and  $l_{ij}=1$ if  $y_i$ and $y_j$ are  known to be dissimilar.  Notice that the similarity indicated by $l_{ij}$ is usually from extra resource instead of the knowledge that can be learned from data instances $y_1, y_2, \cdots,  y_n$ directly.  DRLIM learns a parametric mapping $$A: y \in \mathbb R^K \mapsto z \in\mathbb R^D,$$ such that the embeddings of similar instances attract each other in the low-dimensional space while the embeddings of  dissimilar instances push each other away in the low-dimensional space. In this spirit, the exact loss function of DRLIM is as follows:
\begin{align}
L(A)=&\sum_{(i,j)\in I} (1-l_{ij})\frac{1}{2}\|A(y_i)-A(y_j)\|^2  \label{drlim-loss} \\
&+l_{ij}\frac{1}{2} (max(0, \beta-\|A(y_i)-A(y_j)\| )^2, \notag
\end{align}
where $\beta>0$ is the parameter for the contrastive loss term which decides the extent to which we want to push the dissimilar pairs apart.
Since the parametric mapping $A$ is assumed to be decided by some parameter. DRLIM learn the mapping $A$  by minimizing the loss function in \eqref{drlim-loss} with respect to the parameters of $A$. The mapping A could be either linear or nonlinear.  For example, we can assume $A$ is a two-layer fully connected neural network and then minimize the loss function \eqref{drlim-loss} with respect to the weight. Finally, for any new data instance $y_{new}$, its low-dimensional embedding is represented by $A(y_{new})$  without knowing its relationship to the training instances.

\section{Deep sparse learning framework} \label{dsc}
\subsection{Overview}
\begin{figure}[t]
\centerline{\includegraphics[width=3.6in]{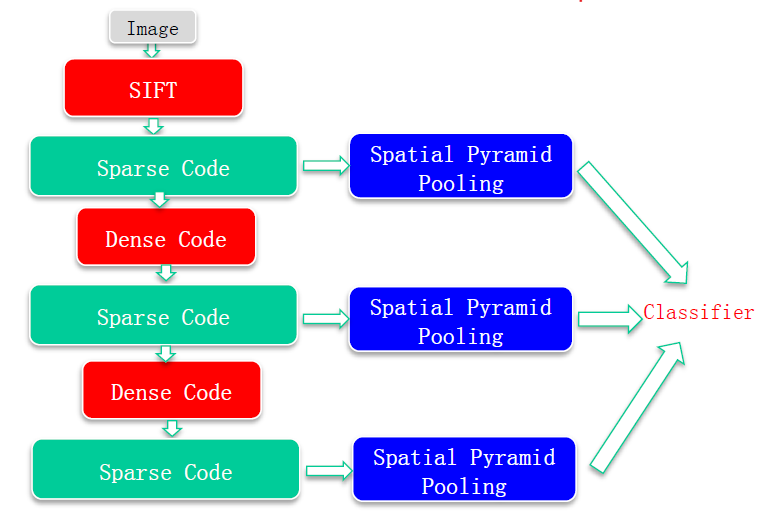}}
\caption{A three-layer deep sparse coding framework. Each of the three layers contains three modules. The first module converts the input (image patches at the the first layer and sparse codes at other layers) to dense codes. The second module is a sparse encoder converting the dense codes to sparse codes. The sparse codes are then sent to the next layer, and simultaneously to a spatial pyramid pooling module. The outputs of the spatial pyramid pooling modules can be used for further tasks such as classification. }\label{arch}
\end{figure}

Recent progress in deep learning~\cite{bengio2012representation} has shown that the multi-layer architecture of deep learning system, such as that of deep belief networks, is helpful for learning feature hierarchies from data, where different layers of feature extractors are able to learn feature representations of different scopes. This results in more effective representations of data and benefits a lot of further tasks.  The rich representation power of deep learning methods motivate us to combine deep learning with the bag-of-visual-words pipeline to achieve better performance on object recognition tasks. In this section, we introduce a new learning framework, named as deep sparse coding (DeepSC), which is built of multiple layers of sparse coding. 


Before we introduce the details of the DeepSC framework, we first identify two difficulties in designing such a  multi-layer sparse coding architecture. 
\begin{itemize}
\item First of all, to build the feature hierarchies from bottom-level features, it is important to take advantage of the spatial information of image patches such that a higher-level feature is a composition of lower-level features. However, this issue is hardly addressed by simply stacking sparse encoders.
\item Second,  it is well-known (see~\cite{wang2010locality,gao2010local}) that sparse coding is not ``smooth'', which means a small variation in the original space might lead to a huge difference in the code space. For instance, if two overlapped image patches have similar SIFT descriptors, their corresponding sparse codes can be very different. If another sparse encoder were applied to the two sparse codes, they would lost the affinity which was available in the SIFT descriptor stage. Therefore, stacking sparse encoders would only make the dimensionality of the feature higher and higher without gaining new informations. 
\end{itemize}

Based on the two observations above, we propose the deep sparse coding (DeepSC) framework as follows. The first layer of DeepSC framework is exactly the same as the bag-of-visual-words pipeline introduced in Subsection \ref{sc}. Then in each of the following layer of the framework, there is a sparse-to-dense module which converts the sparse codes obtained from the last layer to dense codes, which is then followed by a sparse coding module. The output sparse code of the sparse coding module is the input of the next layer. Furthermore, the spatial pyramid pooling step is conducted at every layer such that the sparse codes of current layer are converted to a single feature vector for that layer. Finally, we concatenate the feature vectors from all layers as the input to the classifier. We summarize the DeepSC framework in Figure \ref{arch}. It is important to emphasis that the whole framework is unsupervised until the final classifier.

The sparse-to-dense module is the key innovation of the DeepSC framework, where a ``pooling function'' is proposed to tackle the aforementioned two concerns. The pooling function is the composition of a local spatial pooling step and a low-dimensional embedding step, which are introduced in Subsection \ref{pool-learn} and Subsection \ref{sec-dr} respectively. On one hand, the local spatial pooling step ensures the higher-level features are learned from a collection of nearby lower-level features and hence exhibit larger scopes. On the other hand, the low-dimensional embedding process is designed to take into account the spatial affinities between neighboring image patches such that the spatial smoothness information is not lost during the dimension reduction process.  As the combination of the two steps,  the pooling function fills the gaps between the sparse coding modules, such that the power of sparse coding and spatial pyramid pooling can be fully expressed in a multi-layer fashion.


\subsection{Learning the pooling function} \label{pool-learn}
In this subsection, we introduce the details of designing the local spatial pooling step, which performs as the first part of the pooling function. First of all, we define the pooling function as a map from a set of sparse codes on a sampling grid to a set of dense codes on a new sampling grid. Assume that $G$ is the sampling grid that includes $M$ sampling points on a image, where the any two adjacent sampling points have fixed spacing (number of pixels) between them.  As introduced in Subsection \ref{sc}, each sampling point corresponds to the center of a image patch.  Let $Y=[y_1,\cdots, y_{M}] \in \mathbb R^{K \times M}$ be the sparse codes on the sampling grid $G$, where each $y_i$ is associated with a sampling point on $G$ according to its associated image patch. Mathematically, the pooling function is defined as the map: $$f: (Y, G) \mapsto (Z, G'),$$ where $G'$ is the new sampling grid with $M'$ sampling points and $Z=[z_1,\cdots, z_{M'}] \in \mathbb R^{D \times M'}$ stores the $D$-dimensional dense codes ($D<K$ \footnote{For simplicity, we let $D$ be the same as the dimensionality of SIFT features.}) associated with the sampling points on the new sampling grid $G'$.

\begin{figure}[!]
\centerline{\includegraphics[width=2.1in]{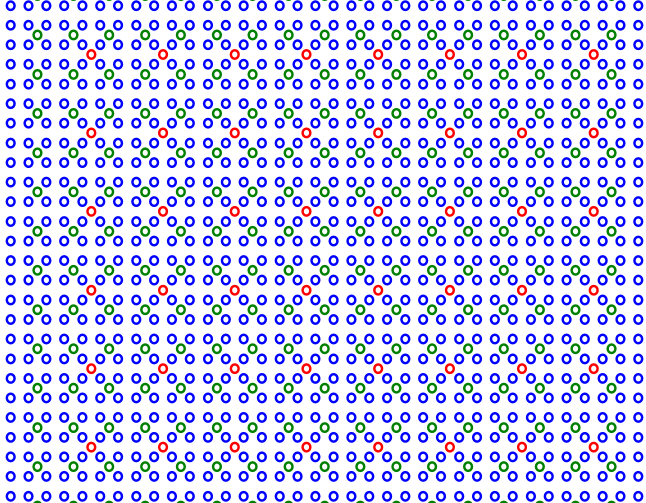}}
\caption{The first, second and third level sampling grids are consists of sampling points in blue, green and red colors, respectively.  The local spatial pooling step is performed on the local $4 \times 4$ grid.}\label{grid}
\end{figure}

As the feature representations learned in the new layer are expected have larger scope than those in the previous layer, we enforce each of the sampling points on new grid $G'$ to cover a larger area in the image. To achieve this, we take the center of  $4\times 4$ neighboring sampling points in $G$ and let it be the new sampling points in $G'$.
By taking the center of every other $4\times 4$ neighboring sampling points, the spacing between neighboring sampling points in $G'$ is twice of that in $G$.
As a result, we map $G$ to a coarser grid $G'$ such that $M' \approx M/4$ (see Figure \ref{grid}).

Once the new sampling grid $G'$ is determined, we finish the local spatial pooling step by applying the max-pooling operator (defined in \eqref{max}) to the subsets of $M$ sparse codes $\{ y_1,\cdots, y_{M}\}$ and obtain $M'$ pooled sparse codes associated with the new sampling grid $G'$. More specifically,  let $\bar{y}_i$ denote the pooled sparse codes associated with the $i$-th sampling point in $ G'$, where $i \in \{1, \cdots, M'\}$. We have 
\begin{equation}
\bar{y}_i:=op_{max}(y_{i_1},y_{i_2}, \cdots, y_{i_{16}}), \label{curr}
\end{equation}
 where $\{{i_1},{i_2}, \cdots, {i_{16}}\}$ are the indices of the $16$ sampling points in $G$ that are most close to the $i$-th sampling point in $ G'$.

\subsection{Dimensionality reduction with spatial information}\label{sec-dr}
In this subsection, we introduce the details of combining the DRLIM method~\cite{hadsell2006dimensionality} with the spatial information of image patches to learn a low-dimensional embedding $A$ such that
\begin{equation}\label{trans}
 z_i:=A(\bar y_i).
\end{equation} As the feature vector is transformed by $A$ to lower-dimensional space, part of its information is discarded while some is preserved.  
As introduced in Subsection \ref{drlim}, DRLIM is trained on a collection of data instance pairs $(\bar y_i, \bar y_j)$, each of which is associated with a binary label indicating their relationship. Therefore, it provides the option to incorporate prior knowledge in the dimensionality reduction process by determining the binary labels of training pairs based on the prior knowledge. 

In the case of object recognition, the prior knowledge that we want to impose on the system is that if a image patch is shifted by a few pixels, it still contains the same object.
Therefore, we constructed the collection of training pairs for DRLIM as follows. We extract training pairs such that there always exist overlapped pixels between the two corresponding patches. Let $\bar y_i$ and $\bar y_j$ be the pooled sparse codes corresponding to two image patches that have overlapped pixels and $d_{ij}$ be the distance (in terms of pixels) between them, which is calculated based on the coordinate of the image patch centers. Given a thresholding $\sigma$, we set
\begin{equation} \label{sigma} l_{ij} = \left\{ 
  \begin{array}{l l}
    0 & \quad d_{ij}<\sigma\\
    1 & \quad d_{ij}>\sigma
  \end{array} \right.
  \end{equation}
 Generated this way,  $l_{ij}=0$ indicates the two image patches are mostly overlapped, while $l_{ij}=1$ indicates that the two image patch are only partially overlapped.  This process of generating training pairs ensures that the training of the transformation $A$ is focused on the most difficult pairs. Experiments shows that if we instead take the pooled sparse codes of far-apart image patches as the negative pairs ($l_{ij}=1$), DRLIM suffers downgrading in performance. The sensitivity of the system to the thresholding parameter $\sigma$ is demonstrated in Table \ref{tab:drlim}.

Let the linear transformation $A$ be defined by the transformation matrix $W \in \mathbb R^{D\times K}$ such that  $$A(\bar y_i) =W\bar y_i,$$ and then the loss function with respect to the pair $(\bar y_i, \bar y_j)$ is
\begin{align}\label{ijloss}
L_{ij}(W) &=(1-l_{ij})\frac{1}{2}\|W\bar y_i-W\bar y_j \|^2 \\
&+ l_{ij} max(0, \beta- \|W\bar y_i-W\bar y_j \|)^2.
\notag
\end{align}

Let $I$ be the set of index pairs for training pairs collected from all training images, $W$ is then obtained by minimizing the loss with respect to all training pairs, i.e., solving
\begin{align}
&\min_W  \sum_{(i,j)\in I} L_{ij} \notag \\
&s.t. \ \ \  \|w_k\|\le 1, \ \ \forall k=1, \cdots, K. \notag
\end{align}

\section{Experiments}
In this section, we evaluate the performance of DeepSC framework for image classification on three data sets: 
Caltech-101 \cite{feifei2004learning} ,  Caltech-256 \cite{griffin2007caltech} and 15-Scene.  Caltech-101 data set contains $9144$ images belonging to $101$ classes, with about $40$ to $800$   images per class. Most images of Caltech-101 are with medium resolution, i.e., about $300 \times 300$. 
 Caltech-256  data set  contains $29,780$ images from 256 categories. The collection has
  higher intra-class variability and object location variability than
  Caltech-101. The images are of similar size to Caltech-101. 
15-Scene data set is compiled by several
  researchers \cite{feifei2005bayesian, lazebnik2006beyond,
    oliva2001modeling},  contains a total of 4485 images falling into
  15 categories, with the number of images per category ranging from
  200 to 400. The categories include living room, bedroom, kitchen, highway, mountain, street and et al.

For each data set, the average per-class recognition accuracy is reported. Each reported number is the average of 10 repeated evaluations with random selected training and testing images.  For each image, following \cite{boureau2010learning}, we sample $16 \times 16$ image patches with 4-pixel spacing  and use $128$ dimensional SIFT feature as the basic dense feature descriptors. The final step of classification is performed using one-vs-all SVM through LibSVM toolkit~\cite{chang2011libsvm}.  The parameters of DRLIM and the parameter to control sparsity in the sparse coding are selected layer by layer through cross-validation. In the following, we present a comprehensive set of experimental results, and discuss  the influence of each of the parameters independently. In the rest of this paper, DeepSC-2 indicates two-layer DeepSC system; DeepSC-3 represents three-layer DeepSC system, and SPM-SC means the one layer baseline, i.e. the BoV pipeline with sparse coding plus spatial pyramid pooling.

\subsection{Effects of Number of DeepSC Layers}
As shown in Figure~2, the DeepSC framework utilizes multiple-layers of
feature abstraction to get a better representation for images. Here we
first check the effect of varying the number of layers utilized in our
framework. Table~\ref{tab:multi} shows the 
average per-class recognition accuracy on three data sets when all
using 1024 as dictionary size. 
The  number of training images per class for the three data sets is set as $30$
  for Caltech-101, $60$ for Caltech-256, and $100$ for 15-Scene
  respectively. The second row shows the results when we have only one
  layer of the sparse coding, while the third row and the fourth row
  describe the results when we have two layers in DeepSC or three
  layers in DeepSC. Clearly the multi-layer structured DeepSC framework has superior performance on all three data sets compared to the single-layer SPM-SC system. Moreover, the classification accuracy improves as the number of layers increases.

\begin{table}[h]
\centering
\begin{tabular}{|c|c|c|c|}
\hline
 & Caltech-101 & Caltech-256 & 15-Scene \\
\hline\hline
SPM-SC &$75.66 \!\pm\! 0.59$  & $43.04 \!\pm\! 0.34$& $80.83 \!\pm\! 0.59$\\
\hline
DeepSC-2 & $77.41 \!\pm\! 1.06$ & $46.02 \!\pm\! 0.57$ &$82.57 \!\pm\! 0.72$ \\
\hline
DeepSC-3 & $78.24 \!\pm\! 0.76$ & $47.00 \!\pm\! 0.45$ & $82.71 \!\pm\! 0.68$\\
\hline
\end{tabular}
\caption{Average per-class recognition accuracy (shown as percentage) on three data sets using 1024 as dictionary size. The
  number of training images per class for the three data sets are $30$
  for Caltech-101, $60$ for Caltech-256, and $100$ for 15-Scene
  respectively. DeepSC-2/3: two/three layers of deep sparse
  coding. SPM-SC: the normal BoV pipeline with one layer of sparse
coding plus spatial pyramid pooling. \label{tab:multi}}
\end{table}

\subsection{Effects of SC Dictionary Size}

We examine how performance of the proposed DeepSC framework changes when varying the dictionary size of the sparse coding.   On each of the three data sets, we consider three settings where the dimension of the sparse codes $K$ is $1024, 2048$ and $4096$. The  number of training images per class for these experiments is set as $30$  for Caltech-101, $60$ for Caltech-256, and $100$ for 15-Scene respectively. We report the results for the three data sets in Table~\ref{tab:dict1},
  Table~\ref{tab:dict2} and Table~\ref{tab:dict3} respectively.  Clearly, when increasing the dictionary size of sparse coding $K$ from 1024 to 4096, the accuracy of the system improves for all three data sets. We can observe that the performance of DeepSC is always improved with more layers, while in the case of $K=4096$ the performance boost in term of accuracy is not so significant. This probably is due to that the parameter space in this case is already very large for the limited training data size. Another observation we made from Table~\ref{tab:dict1}, Table~\ref{tab:dict2} and Table~\ref{tab:dict3} is that DeepSC-2 (K=1024) always performs better than SPM-SC (K=2048), and  DeepSC-2 (K=2048) always performs better than SPM-SC (K=4096). These two comparisons demonstrate that simply increasing the dimension of sparse codes doesn't give the same performance boost as increasing the number of layers, and therefore DeepSC framework indeed benefits from the feature hierarchies learned from the image.
  
 \begin{table}[h]
\centering
\begin{tabular}{|c|c|c|c|}
\hline
Caltech-101 & $K \!= \!1024$ & $K \!=\!2048$ & $K\!=\!4096$ \\
\hline\hline
SPM-SC &$75.66 \!\pm \!0.59$  & $76.34 \!\pm \!0.58$& $77.21 \!\pm\! 0.7$\\
\hline
DeepSC-2 & $77.41 \!\pm \!1.06$ & $78.27 \!\pm \!0.6$ &$78.3 \!\pm \! 0.9$ \\
\hline
DeepSC-3 & $78.24 \!\pm \! 0.76$ & $78.43 \!\pm \! 0.72$ & $78.41 \!\pm \! 0.74$\\
\hline
\end{tabular}
\caption{Effect of dictionary size used in sparse coding on
  recognition accuracy (shown as percentage).  data set: Caltech-101;
  number of training images per class: 30 \label{tab:dict1}}
\end{table}

\begin{table}[h]
\centering
\begin{tabular}{|c|c|c|c|}
\hline
Caltech-256 & $K\!=\!1024$ & $K\!=\!2048$ & $K\!=\!4096$ \\
\hline\hline
SPM-SC &$43.04 \!\pm\! 0.34$  & $45.66 \!\pm\! 0.53$& $47.8 \!\pm\! 0.63$\\
\hline
DeepSC-2 & $46.02 \!\pm\! 0.57$ & $48.04\!\pm\!0.44$ &$49.29\!\pm\!0.50$ \\
\hline
DeepSC-3 & $47.0 \!\pm\! 0.45$ & $48.85\!\pm\!0.42$ & $49.91\!\pm\!0.39$\\
\hline
\end{tabular}
\caption{Effect of dictionary size used in sparse coding on
  recognition accuracy (shown as percentage). data set: Caltech-256;
  number of training images per class: 60  \label{tab:dict2}}
\end{table}

\begin{table}[h]
\centering
\begin{tabular}{|c|c|c|c|}
\hline
15-Scene& $K = 1024$ & $K=2048$ & $K=4096$ \\
\hline\hline
SPM-SC &$80.83 \!\pm\! 0.59$  & $82.11 \!\pm\! 0.61$& $82.88 \!\pm\! 0.82$\\
\hline
DeepSC-2 & $82.57 \!\pm\! 0.72$ & $83.58 \!\pm\! 0.71$ &$83.76 \!\pm\! 0.72$ \\
\hline
DeepSC-3 & $82.71 \!\pm\! 0.68$ & $83.58 \!\pm\! 0.61$ & $83.8 \!\pm\! 0.73$\\
\hline
\end{tabular}
\caption{Effect of varying sparse coding dictionary size on
  recognition accuracy (shown as percentage). data set: 15-Scene;
  number of training images per class: 100  \label{tab:dict3}}
\end{table}

\subsection{Effects of Varying Training Set Size}

Furthermore,  we check the performance change when varying the number of training images per
class on two Caltech data sets. Here we fix the
dimension of the sparse codes $K$ as 2048. 
On Caltech-101, we compare two cases: randomly select $15$ or $30$ images per category
respectively as training images and test on the rest.  
On Caltech-256, we randomly select $60$, $30$ and $15$ images per
  category respectively as training images and test on the
  rest. Table~\ref{tab:trs1} and Table~\ref{tab:trs2} show that with
  the smaller set of training images, DeepSC framework still
  continues to improve the accuracy with more layers.

\begin{table}[h]
\centering
\begin{tabular}{|c|c|c|}
\hline
Caltech-101 & $30$ & $15$ \\
\hline\hline
SPM-SC & $76.34 \!\pm \!0.58$  & $69.94\!\pm \!0.61$\\
\hline
DeepSC-2 & $78.27 \!\pm \!0.6$ & $71.53\!\pm \!0.53$ \\
\hline
DeepSC-3 & $78.43 \!\pm \! 0.72$ & $71.86\!\pm \!0.55$ \\
\hline
\end{tabular}
\caption{Effect of varying training set size on averaged recognition
  accuracy. data set: Caltech-101; Dictionary Size: 2048 \label{tab:trs1}}
\end{table}

\begin{table}[h]
\centering
\begin{tabular}{|c|c|c|c|}
\hline
Caltech-256 & $60$ & $30$ & $15$ \\
\hline\hline
SPM-SC  &$45.66 \!\pm\! 0.53$  & $39.86 \!\pm\! 0.24$ & $33.44 \!\pm\!0.15$\\
\hline
DeepSC-2 & $48.04 \!\pm\! 0.44$ & $41.86 \!\pm\!0.28$ & $35.10 \!\pm\!0.19$\\
\hline
DeepSC-3 & $48.80\!\pm\! 0.42$ & $42.33 \!\pm\!0.29$ & $35.28 \!\pm\! 0.27$\\
\hline
\end{tabular}
\caption{Effect of varying training set size on averaged  recognition
  accuracy. data set: Caltech-256; Dictionary Size: 2048 \label{tab:trs2}}
\end{table}

\subsection{Effects of varying parameters of DRLIM}

In table~\ref{tab:drlim}, we report the performance variations when tuning the parameters for DRLIM. The parameter $\sigma$ is the threshold for selecting positive and negative training pairs (see \eqref{sigma}) and the parameter $\beta$ in the hinge loss (see \eqref{ijloss}) of DRLIM model is for controlling penalization for negative pairs. We can see that it is important to choose the proper thresholding parameter $\sigma$ such that the transformation learned by DRLIM can differentiate mostly overlapped image pairs and partially overlapped image pairs.

\begin{table}[h]
\centering
\begin{tabular}{|c|c|c|c|c|c|c|}
\hline
$\sigma$ $\backslash$ $\beta$ &1 &2 &3&4&5&6\\
\hline
8 & 76.5 & 77.41 & 77.07& 76.71 & 76.24 & 75.81  \\
\hline
16 & 74.93& 76.55 &76.87 & 76.97 & 76.43 & 75.83 \\
\hline
24 & 73.95 & 75.43 &76.18 & 76.42 & 76.53 & 76.45\\
\hline
\end{tabular}
\caption{The effect of tuning DRLIM parameters on recognition accuracy
 for DeepSC-2. data set: Caltech-101; dictionary size: 1024; the number
 of training images per class: 30. 
\label{tab:drlim}}
\end{table}

\subsection{Comparison with other methods}

We then compare our results with other algorithms in Table~\ref{tab:compare}. 
The most direct baselines \footnote{We are also aware of that some works achieve very high accuracy  based on adaptive pooling step~\cite{feng2011geometric} or multiple-path system that utilizes image patches of multiple sizes~\cite{bo2013multipath}. } for DeepSC to compare are 
the sparse coding plus SPM framework (ScSPM) \cite{yang2009linear}, 
LLC\cite{wang2010locality}, and
SSC\cite{bala2013smooth}. Table~\ref{tab:compare} shows the comparison 
of our DeepSC versus the ScSPM and SSC.  We can see that our results
are comparable to SSC, with a bit lower accuracy on the 15-Scene data
(the std of SSC is much higher than ours). For the LLC method proposed
from \cite{wang2010locality}, it reported to achieve 73.44\% for
Caltech-101 when using $K=2048$ and 47.68\% when using $K=4096$. 
Our DeepSC-3 has achieved 78.43\% for
Caltech-101 when using $K=2048$ and 49.91\% when using $K=4096$.
Overall our system achieves the state-of-the-art performance on all
the three data sets. 
%

\begin{table}[t]
\centering
\begin{tabular}{|c|c|c|c|}
\hline
 & Caltech-101 & Caltech-256 & 15-Scene \\
\hline\hline
ScSPM &$73.2 \!\pm\! 0.54$  & $40.14 \!\pm\! 0.91$& $80.28
\!\pm\! 0.93$\\
\hline
SSC & $77.54 \!\pm\! 2.59$ & $-$ & $84.53\!\pm\! 2.57$\\
\hline
DeepSC-3 & $78.24 \!\pm\! 0.76$ & $47.04 \!\pm\! 0.45$ & $82.71
\!\pm\! 0.68$\\
\hline
\end{tabular}
\caption{Comparison of results with other image recognition
  algorithms: ScSPM\cite{yang2009linear},
  LLC\cite{wang2010locality}, and
  SSC\cite{bala2013smooth}. Dictionary size $K=1024$. 
Number of training images are 30, 60, and 100 for Caltech-101, Caltech-256 and 15-Scene respectively.}
\label{tab:compare}
\end{table}

\vspace{-2mm}

\bibliographystyle{paper}
\bibliography{ref}

\end{document}